\documentclass{article}

\usepackage{microtype}
\usepackage{graphicx}
\usepackage{subfigure}
\usepackage{booktabs} 

\usepackage{hyperref}

\usepackage[accepted]{icml2025}

\usepackage{amsmath}
\usepackage{amssymb}
\usepackage{mathtools}
\usepackage{amsthm}

\usepackage[capitalize,noabbrev]{cleveref}

\usepackage{enumitem}
\usepackage{wrapfig}
\theoremstyle{plain}

\theoremstyle{definition}

\theoremstyle{remark}

\usepackage[utf8]{inputenc}
\usepackage{tcolorbox}

\newtcolorbox{definitionbox}[1][]{
  colback=olive!5,         
  colframe=gray!75!black, 
  fonttitle=\bfseries,    
  boxrule=1pt,            
  arc=2mm,                
  left=2mm,               
  right=2mm,              
  top=2mm,                
  bottom=2mm,             
  boxsep=1mm,             
  nobeforeafter,          
  #1 
}

\usepackage[textsize=tiny]{todonotes}
 
\icmltitlerunning{Standard Neural Computation Alone Is Insufficient for Logical Intelligence}
\begin{document}

\twocolumn[
\icmltitle{Standard Neural Computation Alone Is Insufficient \\ for Logical Intelligence}
 
 
\icmlsetsymbol{equal}{*}

\begin{icmlauthorlist}
\icmlauthor{Youngsung Kim}{yyy}
\end{icmlauthorlist}

\icmlaffiliation{yyy}{Department of Artificial Intelligence, Department of Electrical and Computer Engineering, Inha University, Incheon, Republic of Korea}
 
\icmlcorrespondingauthor{Youngsung Kim}{yskim.ee@gmail.com}

\icmlkeywords{Machine Learning, ICML}

\vskip 0.3in
]



\printAffiliationsAndNotice{}  

 
\begin{abstract}
Neural networks, as currently designed, fall short of achieving true logical intelligence. Modern AI models rely on \textbf{standard neural computation}—inner-product-based transformations and nonlinear activations—to approximate patterns from data. While effective for inductive learning, this architecture lacks the structural guarantees necessary for \textit{deductive inference} and logical consistency. As a result, deep networks struggle with rule-based reasoning, structured generalization, and interpretability without extensive post-hoc modifications.  
This position paper argues that \textbf{standard neural layers must be fundamentally rethought to integrate logical reasoning}. We advocate for \textbf{Logical Neural Units (LNUs)}—modular components that embed differentiable approximations of logical operations (e.g., AND, OR, NOT) directly within neural architectures. We critique existing neurosymbolic approaches, highlight the limitations of standard neural computation for logical inference, and present LNUs as a necessary paradigm shift in AI. Finally, we outline a roadmap for implementation, discussing theoretical foundations, architectural integration, and key challenges for future research.
\end{abstract}

\section{Introduction}
\label{sec:intro}

\textbf{Artificial Intelligence (AI)} aims to emulate human intelligence, yet its development has oscillated between two dominant paradigms: \emph{symbolic AI} and \emph{connectionism}. Since the 1956 Dartmouth workshop, symbolic AI has emphasized logical rules and structured inference, whereas connectionist models, such as neural networks, have focused on learning from data \cite{McCulloch1943, Kautz2022}. Modern large-scale neural models trained on symbolic data, such as natural language, can appear to perform deductive reasoning, but they do not robustly replicate human-like logical inference \cite{tian-etal-2021-diagnosing}. The challenge lies in their reliance on statistical approximations rather than explicit rule-based deduction \cite{Hoelldobler1994, BH2005, SEDA2006109, serafini2016logic}.

Symbolic AI, rooted in formal logic and computational rules, enables structured reasoning and systematic problem-solving. However, it faces the \textit{symbol grounding problem}—symbols remain abstract and lack intrinsic meaning unless anchored to real-world sensory experiences \cite{Harnad1990grounding, Barsalou1999perceptualsymbol}. In contrast, connectionist models excel at pattern recognition from raw data but struggle with systematic reasoning, logical rule generalization, and interpretability \cite{McClelland1986, Rumelhart1986}. These complementary strengths and weaknesses suggest that neither approach alone is sufficient for achieving robust, human-like reasoning.

Neuro-symbolic AI seeks to bridge the gap between connectionist learning and symbolic computation by embedding structured reasoning within neural architectures \cite{Hilario1995, Hatzilygeroudis2000, garcez2023wave}. While symbolic theories suggest that structured formal systems provide meaning \cite{Dennett1969, NewellSimon1972, Fodor1975, Haugeland1985}, perceptual symbol theories argue for cognition grounded in sensory-motor experience \cite{Barsalou1999perceptualsymbol}. This distinction extends to AI methodologies, where symbolic systems provide formal inference but lack the adaptability of neural models, while neural networks learn representations flexibly but lack explicit symbolic reasoning \cite{garcez2019neural}.

Recent neuro-symbolic approaches embed symbolic inference within deep learning architectures or integrate learned representations into reasoning engines \cite{wang2019satnet, yang2020neurasp, manhaeve2018deepproblog, garcez2019neural, besold2017neural, yang2019learning, BH2005}, yet aligning continuous neural representations with discrete logic remains a challenge. Large language models (LLMs) employ structured reasoning techniques like chain-of-thought prompting \cite{wei2022chain}, tree-based reasoning \cite{yao2023tree}, and graph-based inference \cite{cai2023large} to enhance multi-step reasoning \cite{sun-etal-2024-determlr}. However, their probabilistic nature causes hallucinations, contradictions, and imprecise fact retrieval, limiting their reliability for strict logical inference \cite{bubeck2023sparks, pmlr-v235-kambhampati24a}. Despite approximating logical reasoning, LLMs lack formal guarantees for symbolic inference, highlighting the need for architectures that embed explicit logical structures to bridge statistical learning and rule-based reasoning \cite{pmlr-v235-kambhampati24a, rae2021scaling, creswell2022selection}.

This paper argues that \textbf{purely arithmetic-based neural layers—those relying on inner products and nonlinear activations—are insufficient for robust logical intelligence}. While neural networks excel at function approximation through inductive learning, they lack inherent support for deductive reasoning and symbolic consistency. Building on early AI logic units and recent neuro-symbolic advances, we propose \emph{Logical Neural Units} (LNUs), which embed differentiable approximations of logical operations (e.g., fuzzy AND/OR) directly within neural layers. Unlike external symbolic engines or loosely integrated reasoning modules, LNUs seamlessly incorporate logical inference into the sub-symbolic learning process, unifying structured reasoning with modern deep learning architectures.

\textbf{Structure.} We position our argument within existing research by first reviewing human intelligence and learning strategies (Section~\ref{sec:rethinking}). We then examine classical, real-valued, and fuzzy logic approaches to illustrate how logical operations extend beyond Boolean domains (Section~\ref{sec:logic-operations}). Next, we revisit universal approximation theorems, highlighting why neural networks, despite their expressive power, do not inherently enable systematic logical reasoning (Section~\ref{sec:logic-and-approx}). A survey of neurosymbolic frameworks (Section~\ref{sec:ns-approaches}) contextualizes the role of LNUs, which we introduce and illustrate through a toy example (Section~\ref{sec:lnu-positioning}). Finally, we discuss open challenges and future directions for integrating logical reasoning into neural networks (Section~\ref{sec:discussion_conclusion}).

\section{Foundations of Logical Reasoning in Neural Systems}
\label{sec:rethinking}
\textbf{Human Intelligence} has been studied throughout history, yet no single, universally accepted definition exists, despite extensive efforts to characterize it through various intellectual frameworks~\cite{Minsky1961, Newell1965}. Since Aristotle, \textit{logic} has been regarded as a fundamental tool for scientific inquiry, often referred to as the \textit{Organon}~\cite{aristotle_organon}. Rooted in symbolic reasoning~\cite{mccarthy_hayes}, logic contrasts with perceptual learning, which does not require explicit symbolic knowledge and is acquired through experience~\cite{Newell1956LTM, whitehead_principia, tarski_semantics}. Humans develop logical reasoning abilities even without formal training, often through structured learning and knowledge transfer.

Cognitive science distinguishes two complementary domains of intelligence~\cite{Cattell1943, Cattell1963}: fluid intelligence, which enables rapid pattern recognition and adaptability, and {crystallized intelligence, which relies on accumulated knowledge and rule-based reasoning. Together, these support both flexible learning and structured problem-solving.

To achieve intelligence, modern AI systems primarily rely on \textbf{inductive learning}, which generalizes patterns from data, enabling adaptation to new scenarios. Neural networks exemplify this approach by identifying statistical associations without predefined rules. Techniques such as statistical machine learning and inductive logic programming (ILP) derive general rules from examples. In contrast, \textbf{deductive reasoning} applies established axioms to derive conclusions with certainty. Systems such as theorem provers ensure logical soundness but lack adaptability. While deduction does not inherently involve learning, it can simulate knowledge accumulation by deriving and storing new facts~\cite{mccarthy_hayes, Newell1956LTM}.

Hybrid approaches in neurosymbolic AI~\cite{BH2005, besold2017neural, garcez2023wave, MARRA2024104062, pami_survey_neSy_2025} integrate inductive learning’s adaptability with deductive reasoning’s precision. Induction discovers patterns, while deduction ensures their logical validity~\cite{nilsson_ai_principles, russell_norvig}. This fusion enhances \emph{flexibility} through data-driven learning, \emph{soundness} via rule-based inference, and \emph{interpretability} by leveraging explicit logical rules.

Logical reasoning unites deduction and induction, enabling structured thought akin to human \textit{System 2} reasoning~\cite{kahneman2011thinking}. Logical operations such as AND, OR, and NOT are fundamental to compositional and hierarchical inference. While neural networks can approximate these operations \cite{McCulloch1943, Newell1956LTM, besold2017neural}, they lack explicit logical representations unless specifically designed to integrate them.

\section{Extending Logical Operators Beyond Classical Rules}
\label{sec:logic-operations}

Classical symbolic AI relies on propositional and predicate logic for knowledge representation \cite{Newell1956LTM, mccarthy_hayes, BH2005}. Propositional logic operates on Boolean variables ($0$ or $1$) and applies logical operators such as AND, OR, NOT, and implication to derive conclusions. However, strict Boolean operations often fall short when dealing with real-world scenarios that involve uncertainty, continuous-valued information, or partial truths. Such rigid structures limit expressiveness, making it difficult to integrate symbolic logic with modern neural models that inherently operate over continuous feature spaces.

To address these limitations, several extensions to classical logic have been developed. Real-valued logic generalizes Boolean truth values by extending them to the interval $[0,1]$, allowing for continuous operators such as t-norms and t-conorms \cite{menger1942statistical, schweizer1961associative, schweizer1983metric, klement2000triangular}. These operators provide smooth approximations of AND and OR functions, enabling reasoning over continuous-valued domains \cite{tarski_semantics, kleene1952metamathematics}. Similarly, fuzzy logic introduces the concept of partial membership, where truth values represent degrees of belonging rather than strict binary assignments. For instance, the concept of ``hot" may be assigned a membership value of $0.7$ rather than being forced into a rigid true or false classification. Logical operations in fuzzy logic, including min, max, and product t-norms, allow for approximate AND and OR computations, making the system more robust to vagueness and imprecise data \cite{zadeh1965fuzzy, dubois1980fuzzy, hajek1998metamathematics}.

Another critical extension is many-valued logic, such as G\"odel and \L ukasiewicz logic, which allows discrete or continuous truth values beyond $\{0,1\}$. These frameworks support reasoning over intermediate truth levels, making them particularly useful for modeling uncertainty and graded reasoning \cite{rescher1969many, smith1988multiple, lukasiewicz1920threevalued, cignoli2013algebraic}. Probabilistic logic further extends classical reasoning by assigning probabilities to logical statements, allowing AI systems to quantify uncertainty rather than relying strictly on binary logic \cite{pearl1988probabilistic, nilsson1986probabilistic, fagin1990uncertainty}. This probabilistic approach has been widely adopted in \textit{Statistical Relational Artificial Intelligence (StarAI)}, which integrates logic with probabilistic graphical models to manage structured and uncertain domains \cite{Getoor2007, DeRaedt2020}.

The integration of these extended logical frameworks is central to our position that purely arithmetic-based neural networks are insufficient for robust logical intelligence. Neural activations operate over continuous domains, yet classical logic assumes discrete, well-defined truths. By incorporating differentiable approximations of logical operations, such as those found in real-valued, fuzzy, many-valued, and probabilistic logic, we can bridge the gap between sub-symbolic learning and symbolic reasoning. These extended operators form the basis for Logical Neural Units (LNUs), which embed logical computation within deep neural architectures. Rather than treating logic as an external rule-based system, LNUs integrate flexible logical reasoning directly within neural computations, addressing the expressiveness limitations of traditional neural models. The following sections further develop this argument by analyzing why universal approximation alone does not guarantee logical consistency and by positioning LNUs within the broader landscape of neurosymbolic AI.


\section{Theoretical Limits of Standard Neural Computation for Logic}
\label{sec:logic-and-approx}

The previous section examined real-valued, fuzzy, and many-valued logic as extensions that generalize classical Boolean operators for handling uncertainty and graded reasoning. While these approaches introduce flexibility, their integration into neural networks remains an approximation rather than an exact logical representation. Neural networks, by design, operate over continuous spaces and optimize for statistical associations rather than strict logical inference. This fundamental difference raises the question: \emph{Can standard neural architectures inherently support logical reasoning, or are they fundamentally constrained in their ability to model symbolic computation?}

\subsection{Universal Approximation Theorem (UAT)}

The \textit{Universal Approximation Theorem (UAT)} states that a feedforward neural network with a single hidden layer and a sufficient number of neurons can approximate any continuous function on a compact subset of \(\mathbb{R}^n\) to arbitrary precision, given a suitable activation function~\cite{hornik1989universal}. Formally, for a compact set \( K \subseteq \mathbb{R}^n \) and a continuous function \( f: K \to \mathbb{R} \), for any \(\epsilon > 0\), there exists a neural network \(\phi(x)\) such that:
\begin{equation}
\textstyle \sup_{x \in K} |f(x) - \phi(x)| < \epsilon.
\end{equation}

While UAT ensures that neural networks can approximate any continuous function, it does not guarantee efficient learning, interpretability, or logical consistency~\cite{cybenko1989approximation, hornik1989universal}. Moreover, it assumes that \( f \) is continuous, which is a critical limitation when applied to discrete logical operations.

\subsection{Limitations of UAT for Logical Reasoning}

Logical reasoning tasks, such as theorem proving and symbolic inference, reveal fundamental shortcomings in applying UAT to logic-based computations. The key issue lies in the nature of logical rules, which require exact, discrete outputs. While neural networks can approximate logical operators such as AND, OR, and XOR using continuous activations, these approximations are computationally inefficient and lack the precision necessary for formal inference.

Moreover, logical reasoning demands \textit{deterministic} conclusions. Standard neural networks, optimized via gradient-based learning, introduce approximations that may cause outputs to oscillate near correct logical values without strictly converging. This contrasts with formal symbolic systems, where rule-based computations guarantee correctness. The problem is further exacerbated by logical inference operating over \textit{unbounded domains}, such as universal quantifiers (\(\forall x\)), whereas UAT is constrained to compact subsets of \(\mathbb{R}^n\). This mismatch limits neural networks' ability to ensure \textit{global consistency}, which is crucial for systematic reasoning.

Another significant limitation is \textit{interpretability}. Symbolic reasoning systems produce explicit inference chains that can be traced and verified, while neural networks trained under the UAT framework rely on distributed representations that are inherently opaque. The lack of explicit reasoning steps makes it difficult to validate logical consistency, posing challenges for domains requiring explainability.

\subsection{How UAT Constraints Affect Logical Computation}

Although UAT highlights the expressive power of neural networks, it also exposes fundamental challenges in logical computation. Logical reasoning requires exact, discrete, and symbolic representations, which are not naturally supported by continuous approximations. While inductive reasoning in neural networks can approximate logical functions, it does not guarantee soundness, completeness, or consistency in a formal sense. Consequently, additional mechanisms—such as neurosymbolic models incorporating explicit logical rules (\(\land\), \(\lor\), \(\neg\)) or fuzzy t-norms—are needed to bridge the gap between statistical learning and symbolic reasoning.

\subsection{Motivation for Logic-Based Architectures}

The limitations of UAT for logical reasoning highlight the need for hybrid architectures that integrate symbolic constraints within neural networks. Standard neural models struggle with discrete logical computation, global consistency, and interpretability—key elements of systematic reasoning. By embedding logical structures into neural computations, hybrid models improve consistency and explainability.  

Existing neurosymbolic approaches attempt to bridge neural learning and symbolic inference using hard-coded rules or differentiable approximations of logical operators  \cite{BH2005, pami_survey_neSy_2025}. However, many depend on external reasoning modules, reducing scalability and architectural cohesion. A more integrated solution requires embedding trainable logical operators (e.g., differentiable AND, OR, and NOT) directly within neural layers, allowing for seamless interpolation between symbolic and sub-symbolic reasoning.  

The next section reviews existing neurosymbolic frameworks and introduces a novel extension that enhances logic-based computation within neural architectures.

\section{Shortcomings in Existing Neural Logic Approaches}
\label{sec:ns-approaches}

Neurosymbolic AI combines sub-symbolic learning (e.g., neural networks) with symbolic reasoning (e.g., logic-based inference), aiming to unify the efficiency of neural methods and the interpretability of symbolic systems \cite{Kautz2022, BH2005, nsai_Sarker_2021, pami_survey_neSy_2025}. Existing taxonomies (e.g., \cite{Kautz2022}) categorize these methods by how tightly neural and symbolic components are intertwined, from loosely coupled pipelines to deeply integrated architectures. Here, we focus on four representative frameworks---\emph{Neural Logic Networks (NLNs)}, \emph{Logical Neural Networks (LNNs)}, \emph{Logic Tensor Networks (LTNs)}, and \emph{Neural Logic Machines (NLMs)}---each of which approximates logical operations or embeds logical structures into neural models.

\subsection{Parameterized Logic Representations: Multi-Ary Atoms}
Both NLNs and LNNs seek to generalize classical AND/OR with learnable weights, making logical operations \emph{differentiable} and enabling partial-truth inputs in $[0,1]$. This reveals a shared philosophy of embedding logical structure directly into network computations, thereby preserving some interpretability at the operator level. 

\textbf{Neural Logic Networks (NLNs)}~\cite{yang2019learning} map Boolean operators (e.g., AND, OR, XOR) into \emph{weighted, many-valued} logic, frequently using \textbf{product}-based formulations:
\begin{align}
 \textstyle  y_{\text{AND}} &= \textstyle \prod_{i=1}^n \Bigl[1 - w_i\,(1 - x_i)\Bigr], \\
\textstyle  y_{\text{OR}} &= \textstyle 1 - \prod_{i=1}^n \Bigl[1 - w_i\,x_i\Bigr],    
\end{align}
where $w_i \in [0,1]$ are membership weights. 
This approach can excel at discrete algorithmic tasks (e.g., addition, sorting). However, product-based logic can lead to near-zero outputs when multiple inputs $x_i<1$, potentially undermining stability in high-dimensional or uncertain settings.

\textbf{Logical Neural Networks (LNNs)}~\cite{riegel2020logical}, in contrast, implement AND/OR with weighted \textbf{summations}:
\begin{align}
\textstyle  y_{\text{AND}} &= \textstyle f\!\Bigl(\beta - \!\sum_{i \in I} w_i\,(1 - x_i)\Bigr), \\
\textstyle y_{\text{OR}}  &= \textstyle f\!\Bigl(1 - \beta + \!\sum_{i \in I} w_i\,x_i\Bigr),    
\end{align}
where $x_i\in[0,1]$ are partial-truth inputs, $I$ represents the input set, $w_i \ge 0$ are learnable weights, $\beta\ge0$ is a bias term, and $f(\cdot)$ is typically  a clipping function or a ReLU-like activation ensuring $y \in [0,1]$. 
This \textbf{sum}-based formulation can be more robust than a pure product, but calibrating weights/biases in deeper networks may be complex, and LNNs often focus on higher-level rule constraints rather than layer-by-layer embedding.

\subsection{Neural Approximation of Logical Quantifiers via Perceptrons}
\textbf{Logic Tensor Networks (LTNs)}~\cite{serafini2016logic} embed first-order logic into tensor-based neural systems by mapping logical predicates to continuous embeddings, then applying fuzzy t-norms for AND/OR. They unify multiple tasks (e.g., classification, clustering, relational reasoning) yet can become complex in deep architectures where large numbers of predicates and constraints must be managed.

\textbf{Neural Logic Machines (NLMs)}~\cite{dong2019neural} extend multi-valued logic with iterative layers for multi-hop inference. They use MLP modules to approximate quantifiers ($\forall,\exists$), enabling advanced relational reasoning (e.g., family-tree or grid-world tasks). Although powerful, NLMs can be challenging to scale due to their reliance on standard MLP blocks for quantifier operations, which can hinder full interpretability in large domains.

\section{Introducing Logical Neural Units (LNUs) for Modular and Scalable Logical Reasoning}
\label{sec:lnu-positioning}

Existing neurosymbolic approaches incorporate logical operations in various ways but face several limitations. Some rely on product-based fuzzy logic for Boolean connectives, which can become numerically unstable in high-dimensional settings~\cite{yang2019learning}. Others inherit the constraints of perceptron-based architectures, limiting their expressivity for structured reasoning~\cite{riegel2020logical}. While some methods extend to first-order logic or multi-hop relational quantifiers, they often require additional modules or iterative mechanisms, impacting scalability and interpretability~\cite{serafini2016logic, dong2019neural}.  

\subsection{Core Principles for Logical Neural Units (LNUs)}
These challenges motivate the development of \emph{Logical Neural Units (LNUs)}, a framework that embeds logical operations \emph{directly} into deep networks, ensuring stable and scalable neurosymbolic reasoning. LNUs enable smooth interpolation between soft and hard logic, allowing adaptation across tasks and domains.

Scalability is a key consideration. Product-based logic can suffer from vanishing gradients, while MLP-based quantifiers often lack interpretability. LNUs must balance numerical stability with effective layering in deep networks. A hierarchical design that stacks multiple LNUs while maintaining transparent parameters offers both efficiency and clarity.

LNUs integrate seamlessly with existing architectures. Instead of replacing all neural components, LNUs selectively substitute dense layers while retaining standard MLP or attention-based modules for feature extraction. This hybrid approach maintains compatibility with established architectures such as Transformers while minimizing design overhead.

\subsection{Illustrative LNU Architecture}
LNUs generalize and unify prior differentiable logic methods within a single neural unit. Instead of standard \texttt{Linear}+\texttt{ReLU} layers, we define an LNU block as:
\begin{equation}
\mathrm{LNU}(x_1,\dots,x_n; \theta) =
\Bigl[ T_{\land} (g_1(x); \theta_1), T_{\lor} (g_2(x); \theta_2), \dots \Bigr],
\end{equation}
where \( T_{\land} \) and \( T_{\lor} \) are learnable t-norm and t-conorm operators approximating logical AND and OR functions. The functions \( g_i(x) \) map the input vector \( x = (x_1, \dots, x_n) \) to real-valued logic features. The parameter set \( \theta \) consists of adaptive weights that dynamically adjust logical compositions during training, allowing LNUs to learn task-specific dependencies.

\subsection{LNU Layer Composition}
\label{sec:lnu-composition}

Each input feature, interpreted as a truth degree in \([0,1]\), is combined via \emph{learnable logic} (e.g., softmin, softmax, or polynomial expansions) to approximate logic operations.

\paragraph{Differentiable Logic Approximation.}
Classical Boolean and fuzzy logic operations, such as AND and OR, are often computed using \(\min\) and \(\max\) functions, as in G\"odel T-norm and T-conorm  \cite{hajek1998metamathematics, VANKRIEKEN2022103602}.  To facilitate smooth gradient-based learning, LNUs approximate these operations using softmin and softmax:
\[
\text{softmax}(\beta \mathbf{z})_i  \approx \max(\mathbf{z}),~
\text{softmin}(\beta \mathbf{z})_j  \approx \min(\mathbf{z})~
\text{as } \beta \to \infty.
\]
Here, each input \( z_i \) is defined as a weighted feature:
\begin{equation}
z_i = x_i \cdot w_i, \quad \text{where } x_i \in \mathbf{x},~ w_i \in \mathbf{w},~ i=1,\dots,d.    
\end{equation}
The parameter \(\beta \geq 0\) controls the \textit{sharpness} of the approximation, with larger values making the function more Boolean-like. The softmin function is efficiently computed via:
\( 
\text{softmin}(\beta \mathbf{z}) = \text{softmax}(-\beta \mathbf{z}).
\)

\paragraph{Locally Gated Logical Consistency.}
To ensure logical consistency, each feature’s contribution is weighted via learned importance factors. Given an input vector \(\mathbf{z} = (z_1, \dots, z_d)\), the differentiable logic functions are:
\begin{align}
    \texttt{soft-OR}(\mathbf{z})
&= \textstyle  \sum_{i=1}^{d} \bigl[\text{softmax}(\beta \mathbf{z})\bigr]_i \cdot z_i,\\
    \texttt{soft-AND}(\mathbf{z})
&= \textstyle  \sum_{i=1}^{d} \bigl[\text{softmin}(\beta \mathbf{z})\bigr]_i \cdot z_i.
\end{align}
For high-dimensional inputs, normalizing outputs by \(\sqrt{d}\) mitigates activation scaling issues and ensures numerical stability. Additionally, the negation operator is defined as:
\(
\texttt{soft-NOT}(x) = 1 - x, ~~~ \text{or} ~~~ 1 - \sigma(W_{\mathrm{not}} x),
\)
where \(\sigma\) is the sigmoid function and \(W_{\mathrm{not}}\) is a learnable parameter.

\paragraph{LNU Layer Architecture.}
LNUs replace standard dense layers by embedding differentiable logical operators instead of matrix multiplication and activation functions. Given an input matrix \(\mathbf{X} \in \mathbb{R}^{n \times d}\) and trainable parameter matrices \(\mathbf{W}_{\text{AND}}, \mathbf{W}_{\text{OR}} \in \mathbb{R}^{d \times o}\), we apply broadcasted element-wise multiplication to implement the \textbf{Locally Gated Logical Consistency} mechanism (details in Appendix~\ref{appendix:lnu-details}).

\paragraph{Deep Logical Networks with LNU Stacking.}
Stacking multiple LNU layers enables a deep hierarchical architecture where each layer progressively refines logical inferences at increasing levels of abstraction:
\begin{equation}
\mathbf{X}^{(\ell)} = \mathrm{LNU}(\mathbf{X}^{(\ell-1)}), \quad \ell = 1, \dots, L.    
\end{equation}
To enhance gradient flow and iterative logical refinement, we incorporate logical residual connections:
\begin{equation}
\mathbf{X}_{\mathrm{res}} = \texttt{soft-IMPLY}(\mathbf{X}, F(\mathbf{X})),    
\end{equation}
where \(\texttt{soft-IMPLY}(A, B) = \texttt{soft-OR}(1 - A, B)\), ensuring stable logical propagation across layers. Additionally, other logical connectives can be incorporated using \texttt{soft-AND} and \texttt{soft-OR}, enabling more expressive reasoning within deep architectures.

\begin{figure}[tb]
\centering
\includegraphics[width=0.90\linewidth]{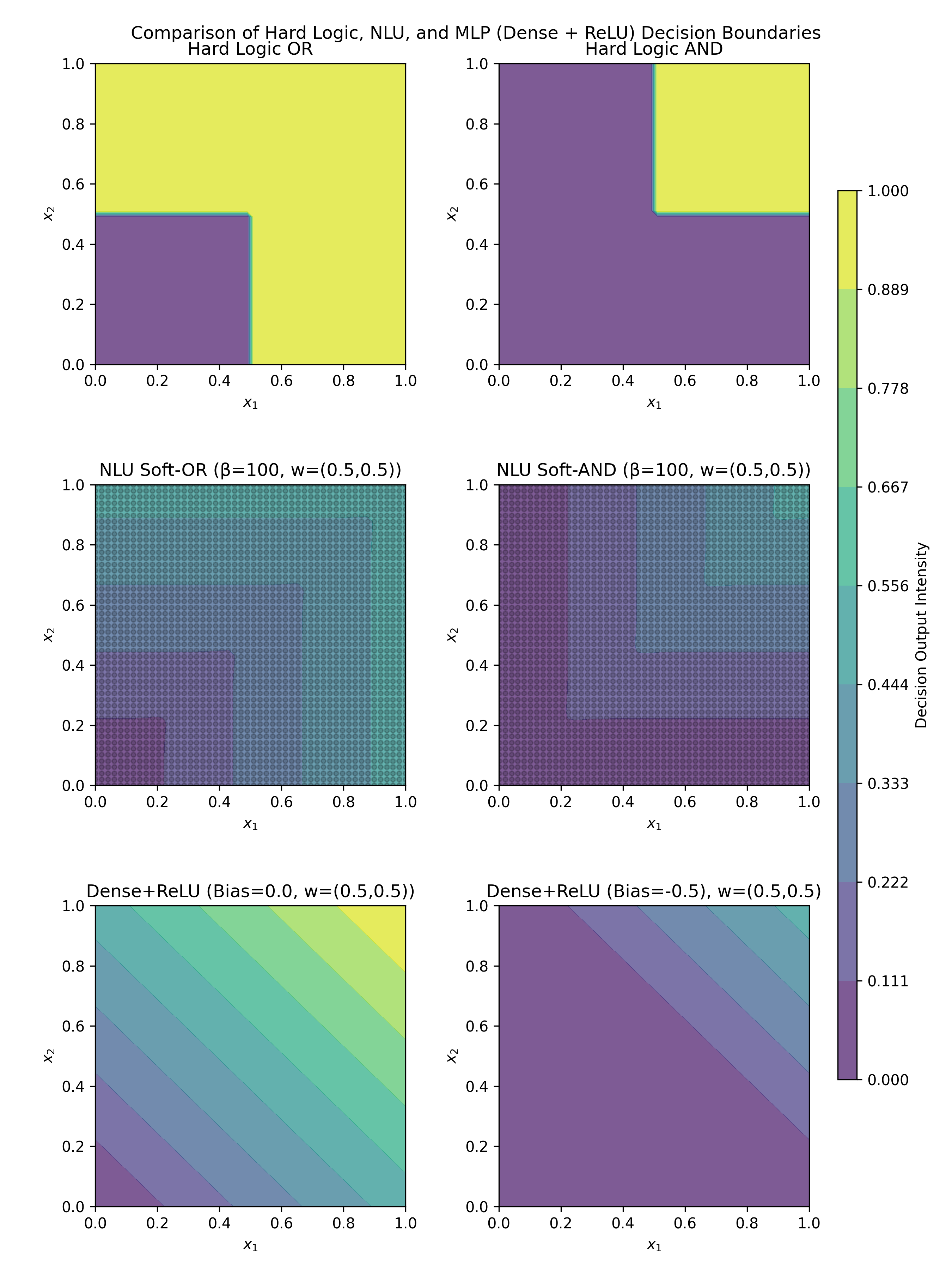}
\vspace{-1.5em}
\caption{Comparison of logical function approximations across different models. The first row represents \textit{Hard Logic}, where AND and OR operations are applied to binarized inputs ($x_i > 0.5, \forall i$). The second row shows LNUs with fixed weights ($w_i=0.5, \forall i$) and a sharp gating parameter $\beta=100$. The third row presents an inner-product unit with ReLU activation, using the same fixed weights ($w_i=0.5, \forall i$) and different biases ($0.0 \text{ or } -0.5$) over the input domain $(x_1,x_2) \in [0,1]$. LNUs closely approximate discrete logical functions, resembling the behavior of hard logic in the first row. Since inputs range over $[0,1]$, LNU outputs remain continuous, allowing for an adjustable decision boundary around the corresponding hard logic outputs. In contrast, the inner-product unit, which relies on summation-based arithmetic, does not exhibit logical function behavior. Additional results for varying values of $\beta$ in LNUs are provided in Appendix~\ref{appx:LNU_beta}.}
\label{fig:lnu_decisionboundary}
\end{figure}

\subsection{Evaluating LNU Interpretability and Decision Boundaries}
\label{sec:interpretability_check} 

Logical operations such as AND and OR follow explicit compositional rules, making their decisions inherently transparent. \textbf{Neural Logic Units (NLUs)} enhance interpretability by structuring decisions through learned logical operations. To assess this interpretability, we analyze decision boundaries and compare them against standard inner-product-based layers.

Visualizing decision boundaries is a common approach to understanding model behavior. Here, we constrain input data to the range \( [0,1] \) and fix the weights for both NLUs and dense (inner-product) layers, ensuring an even distribution that aligns with hard AND/OR logic. This setup allows us to evaluate whether decision-making can be explained at the level of a single computation unit’s output. Specifically, we examine how NLUs approximate hard logic functions such as AND and OR.

As shown in Figure~\ref{fig:lnu_decisionboundary}, NLUs exhibit decision boundaries that closely align with logical structures, whereas inner-product-based layers rely on learned decision surfaces that are harder to interpret. The key advantage of NLUs is their structured decision-making, which inherently follows predefined logical rules, unlike traditional dense layers that require additional interpretation.

When irrelevant features are present, NLUs can filter them out through a gating mechanism, similar to self-attention in a learnable weight setting. This interpretability arises from two key properties: (1) the gating process highlights dominant features, enhancing transparency in the decision process, and (2) predefined logical operations provide structured reasoning, reducing the need for post-hoc interpretation. These properties contribute to NLUs' ability to offer more explainable and structured decision-making compared to conventional neural models.

\subsection{Toy Example: Learning a Logical Function} 
To evaluate the potential of logic-friendly modules, we conduct experiments on a toy logical task. This example showcases the advantages of the LNU Layer in capturing discrete logical patterns, particularly in low-data regimes. 

\paragraph{Toy Logical Functions.}
We test a simple logical expression to assess the model's ability to perform logical inference.

\textbf{Task:} Given the logical function: $f(x_1, x_2, x_3) = (x_1 \lor x_2) \land \lnot(x_3)$, 
the goal is to learn this function from a subset of possible \((x_1, x_2, x_3)\) combinations. We generate random input values in the range \([0,1]\), using 20 samples for training and 200 samples for testing. In data generation, inputs are binarized such that \( x_i > 0.5 \) is mapped to 1, and otherwise 0. This ensures that the target outputs adhere strictly to logical evaluations in the set \(\{0,1\}\).

For comparison, we benchmark Logicron (LNU-based model) against a standard perceptron. Both models consist of a single hidden unit, where Logicron uses an LNU layer and Perceptrons use an inner-product-based linear unit with non-linear activation functions (Sigmoid, ReLU, and GeLU). The output layer in both models consists of a dense layer with a linear transformation followed by a sigmoid activation. These models are referred to as \textit{Logicron} and \textit{Perceptron} in Figure~\ref{fig:lnu_acc}.

\textbf{Observation:} In the toy logical function task, Logicron generalizes more reliably during testing, likely due to its built-in logical gating mechanisms. In contrast, Perceptrons exhibit lower test accuracy, even though all models achieve 100\% training accuracy. 
To ensure a fair comparison, we maintain a similar number of parameters across models: (Perceptron with 97, Logicron with 90, Logicrons +negation unit with 110). At the final epoch (30), the test accuracy (mean $\pm$ standard deviation) for each model is as follows, MLP-Sigmoid: 77.4$\%$ $\pm$1.6, MLP-ReLU : 80.2$\%$ $\pm$1.6, MLP-GeLU : 80.3$\%$ $\pm$1.4, Logicron (softmax) : 84.7$\%$ $\pm$1.7, Logicron neg (softmax): 84.7$\%$ $\pm$1.8.

\begin{figure}[tb]
\begin{center}
\includegraphics[width=0.65\linewidth]{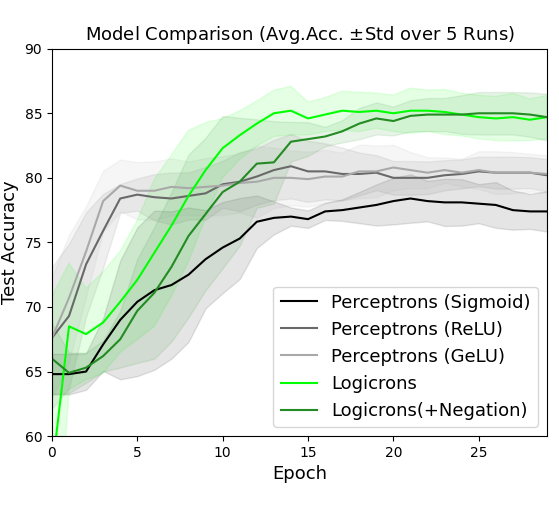}
\end{center}
\vspace{-2em}
\caption{Test accuracy (mean $\pm$ standard deviation, with shading) across epochs for different models, including Perceptrons with various activation functions and Logicrons (LNU layer + dense output layer). Logicron with an additional negation unit is marked as ``+Negation".}
\label{fig:lnu_acc}
\end{figure} 


\section{Discussion and Future Directions}
\label{sec:discussion_conclusion}

\subsection{Position Summary}
We have demonstrated that while standard inner-product-based neural networks excel at sub-symbolic pattern recognition, they inherently lack the capacity for structured logical reasoning. As a result, achieving consistent logical inference often requires large datasets, extensive parameter tuning, and auxiliary learning strategies.

Modern AI models have followed a scaling law, which describes how system performance improves as model and data sizes increase, often following a power-law relationship~\cite{Bahri2024scaling, kaplan2020scaling}. This trend is grounded in the universal approximation theorem, which states that sufficiently large neural networks can approximate a broad range of functions. Consequently, current inductive learning frameworks rely on ever-increasing model size and data volume to capture complex relationships across different domains. However, while scaling enhances function approximation, it does not inherently enable structured, rule-based reasoning.

This paper argues for integrating \textbf{logic-friendly modules} into large-scale neural models to enable implicit logical reasoning. By explicitly incorporating deductive reasoning, these modules enhance logical inference without requiring fundamental architectural changes. We extend existing neurosymbolic frameworks by introducing Logical Neural Units (LNUs), which embed approximate logical operations (e.g., AND, OR, NOT) directly into neural architectures using weighted real-valued logic. These units bridge the gap between sub-symbolic efficiency and symbolic reasoning, providing a scalable alternative to brute-force model expansion. By unifying neural and logical reasoning, we propose a shift away from continuous scaling, which consumes vast computational resources and energy, towards a more structured and efficient approach to AI development.

\subsection{Critical Limitations and Challenges}

LNUs provide a promising bridge between sub-symbolic learning and symbolic reasoning, yet several challenges must be addressed to enhance their effectiveness. One major limitation is their current focus on propositional logic, which restricts their ability to handle more expressive reasoning frameworks. LNUs do not yet incorporate first-order logic constructs such as quantifiers (\(\forall, \exists\)) or domain iteration. Extending LNUs to inductive reasoning frameworks, such as Inductive Logic Programming (ILP) \cite{Muggleton1996} or differentiable logical networks \cite{sen2022neuro}, could significantly improve their ability to capture structured relationships while maintaining scalability.

Another challenge is ensuring logical soundness and completeness. Traditional symbolic reasoning systems guarantee that every derived statement is valid (soundness) and that all logically entailed statements can be derived (completeness). Neural networks, however, rely on approximate optimization and do not inherently provide formal correctness guarantees. While LNUs introduce logical biases into learning, future research should explore differentiable proof-checking mechanisms to enforce logical validity in neural architectures.

The symbol grounding problem remains a fundamental issue in neurosymbolic AI. Learned representations in neural networks often fail to meaningfully align with symbolic concepts, leading to inconsistencies between learned models and human-interpretable logical reasoning. The challenge of mapping abstract symbols to real-world meaning \cite{Harnad1990grounding} necessitates structured constraints or hybrid approaches that integrate both perceptual and logical representations \cite{garcez2023wave, Barsalou1999perceptualsymbol}. Ensuring that real-valued embeddings preserve logical consistency is critical for effective reasoning.

Computational complexity is another challenge. Logical gating mechanisms, such as weighted t-norms with elementwise products, introduce additional overhead, especially in deep architectures. Efficient approximations, such as polynomial expansions as alternatives to softmax-based formulations, can help mitigate these costs while preserving interpretability.

\subsection{Alternative Views}

Despite the advantages of modular neural logic units (e.g., LNUs), alternative perspectives question their necessity. One argument is that sufficiently large MLPs can approximate logical functions without explicitly embedding logical operations. Given sufficient data and computational resources, neural networks can learn complex nonlinear functions that approximate logical inference. However, this approach is inefficient, requiring significantly larger datasets and lacking interpretability. The absence of interpretability makes debugging and improving specific weaknesses in large models challenging. Addressing localized errors may require retraining the entire model, potentially disrupting global reasoning outcomes.

Another perspective suggests that classical fuzzy logic and G{\"o}del’s $\min/\max$ operators already provide logical approximations without modifying network architectures. While these methods capture partial truth values, they lack trainable feature weighting and smooth differentiability, making seamless integration with neural models difficult. Modular neural logic units extend these ideas by embedding logical operations directly into neural computations, allowing for greater adaptability.

Chain-of-thought prompting in transformers has demonstrated that structured reasoning can emerge through large-scale training and carefully designed prompts. However, this remains a heuristic rather than an architectural guarantee of logical consistency and often requires additional optimization during model training. Modular neural logic units complement such approaches by embedding structured logical operations within the network itself, reducing reliance on post-hoc prompt engineering.

Logical constraint learning, as exemplified by SATNet~\cite{wang2019satnet} and neural theorem proving~\cite{rocktaschel2017end}, explicitly encodes logical rules within neural models. While effective, these methods often require manually specified rules, limiting scalability. Modular neural logic units aim to internalize logical inference within neural networks, balancing adaptability with the ability to perform formal reasoning without external supervision.

\subsection{Future Directions}

Several research directions remain for advancing modular neural logic units (e.g., LNUs). One key area is extending LNUs beyond propositional logic by incorporating quantifiers and domain-specific rule induction. This would enable higher-order reasoning, bridging the gap between propositional and first-order logic. Another important step is integrating LNUs with sequence-based models, such as Transformers and RNNs, to enhance structured reasoning in natural language processing and decision-making tasks.

Empirical benchmarking in real-world applications, such as legal reasoning, medical diagnostics, and scientific discovery, is essential to demonstrate the practical effectiveness of LNUs. Additionally, rigorous evaluation frameworks for logical consistency are needed. Establishing soundness and completeness benchmarks for neuro-symbolic reasoning will ensure that LNU-based models adhere to formal logical principles while maintaining the flexibility of neural learning.

\subsection{Concluding Remarks}
This paper has argued that the arithmetic foundations of standard neural networks—inner products and non-linear activations—are insufficient for achieving robust logical intelligence. While large-scale models can approximate logical functions, they fail to ensure logical consistency, interpretability, or rule-based deduction without auxiliary reasoning mechanisms. By seamlessly integrating neural logic units into the core of neural networks, large-scale models may mitigate the challenges associated with scaling while improving their reasoning capabilities.

\section*{Impact Statement}

This paper presents work whose goal is to advance the field of 
Machine Learning. There are many potential societal consequences 
of our work, none which we feel must be specifically highlighted here.

\bibliographystyle{icml2025}
\bibliography{references}

\newpage
\appendix
\onecolumn

\section{Observations on Decision Boundary}\label{appx:LNU_beta}
Expanding on Figure~\ref{fig:lnu_decisionboundary}, we analyze the effect of different gating parameters $\beta \in \{1, 10, 100\}$ on LNU decision boundaries. As $\beta$ increases, the boundaries transition from smooth to sharper logical separations, demonstrating a controllable interpolation between continuous and discrete decision-making within a single neural computation unit, such as AND/OR.

\begin{figure}[t]
\centering
\includegraphics[width=0.4\linewidth]{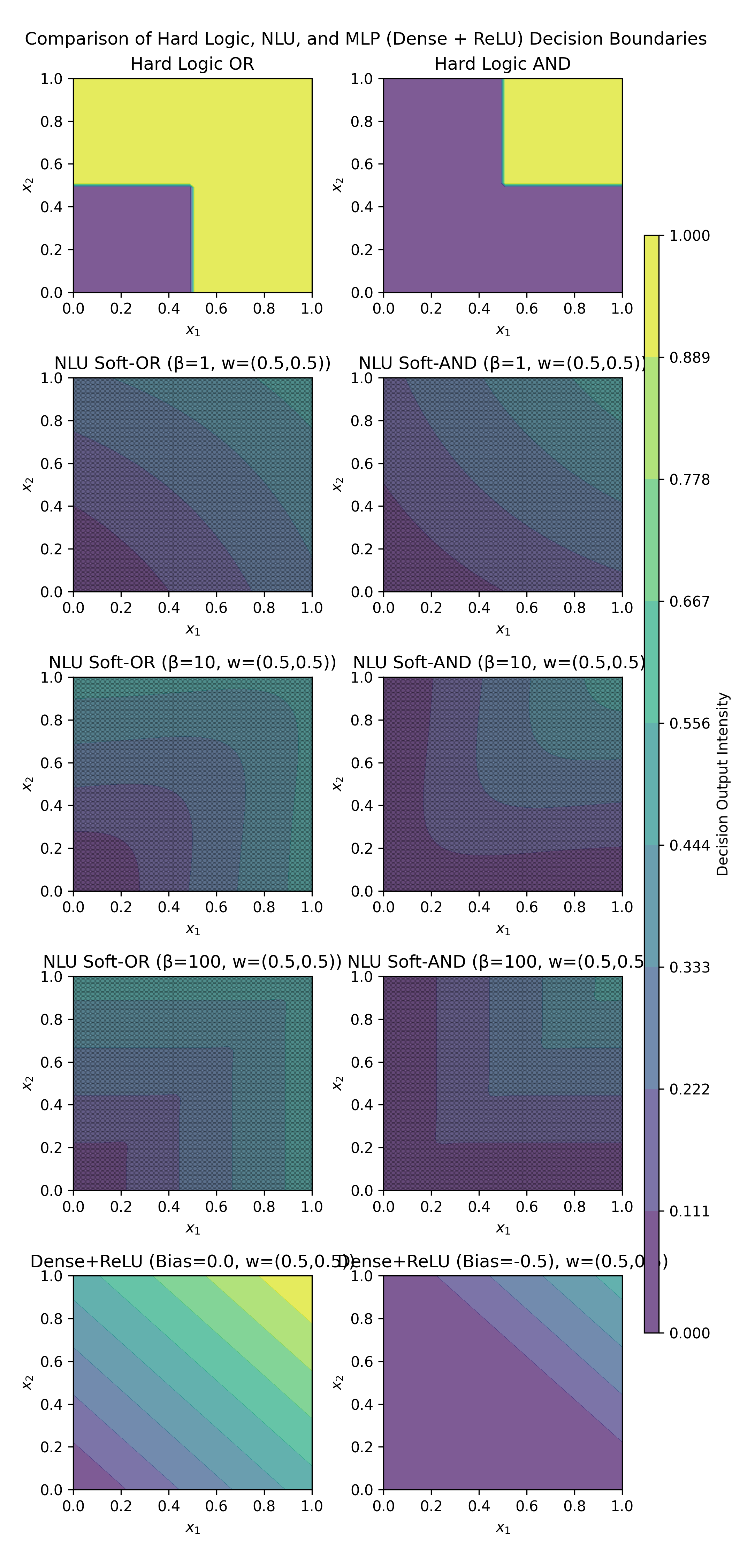}
\vspace{-1.5em}
\caption{Effect of gating parameter $\beta \in \{1, 10, 100\}$ on LNU decision boundaries with fixed weights ($w_i=0.5, \forall i$). Increasing $\beta$ sharpens the boundary, approaching hard logic.}
\label{fig:lnu_decisionboundary_appx}
\end{figure}

\section{Matrix operations}\label{appendix:lnu-details}
Given an input matrix \(\mathbf{X} \in \mathbb{R}^{n \times d}\) and trainable parameter matrices \(\mathbf{W}_{\text{AND}}, \mathbf{W}_{\text{OR}} \in \mathbb{R}^{d \times o}\), we apply broadcasted element-wise multiplication to implement the \textbf{Locally Gated Logical Consistency} mechanism
\[
\mathbf{Z}_{\text{AND}} = \mathbf{X} \odot \mathbf{W}_{\text{AND}}, \quad
\mathbf{Z}_{\text{OR}} = \mathbf{X} \odot \mathbf{W}_{\text{OR}}.
\]
Here, broadcasting ensures:
\[
z_{ijk} = x_{ij} \cdot w_{jk}, \quad x_{ij} \text{ is shared across } k, \quad w_{jk} \text{ is shared across } i.
\]

Logical transformations are then applied to produce the final outputs:
\[
\mathbf{X}_{\mathrm{AND}} = \texttt{soft-AND}(\mathbf{Z}_{\mathrm{AND}}), ~~~ 
\mathbf{X}_{\mathrm{OR}} = \texttt{soft-OR}(\mathbf{Z}_{\mathrm{OR}}).
\]
The final LNU layer output is formed by concatenating these logical transformations:
\[
\mathbf{X}_{\mathrm{out}} = \bigl[\mathbf{X}_{\mathrm{AND}} \;\|\; \mathbf{X}_{\mathrm{OR}}\bigr].
\]
Optionally, \(\texttt{soft-NOT}\) transformations can be included to enhance representational flexibility.

\end{document}